\documentclass[acmsmall]{acmart}

\AtBeginDocument{%
  }

\usepackage{makecell}
\usepackage{multirow}
\usepackage{threeparttable}
\usepackage[noend]{algpseudocode}
\usepackage{algorithm}
\usepackage{graphicx}
\usepackage{booktabs}
\usepackage{amsmath}
\usepackage{amsfonts}
\usepackage{color}
\usepackage{url}

\begin{document}

\title{Token-Event-Role Structure-based Multi-Channel Document-Level Event Extraction
}

\author{Qizhi Wan}
\thanks{This research was partially supported by the National Natural Science Foundation of China (Nos. 61972184, 62272205, 62076112, and 62272206), the Natural Science and Foundation of Jiangxi Province (20212ACB202002), and the Funding Program for Academic and Technical Leaders in Major Disciplines of Jiangxi Province (20213BCJL22041)}
\email{wanqizhi1006@163.com}
\orcid{0000-0002-8835-5134}
\author{Changxuan Wan}
\email{wanchangxuan@263.net}
\orcid{0000-0002-6222-1015}
\affiliation{%
  \institution{Jiangxi Key Laboratory of Data and Knowledge Engineering, China and School of Information Management, Jiangxi University of Finance and Economics}
  \city{Nanchang}
  \state{Jiangxi}
  \country{China}
}

\author{Keli Xiao}
\email{Keli.xiao@stonybrook.edu}
\orcid{0000-0001-6494-1174}
\affiliation{%
  \institution{College of Business, Stony Brook University}
  \city{Stony Brook}
  \state{New York}
  \country{USA}}

\author{Hui Xiong}
\email{xionghui@ust.hk}
\affiliation{%
  \institution{Thrust of Artificial Intelligence, Hong Kong University of Science and Technology (Guangzhou)}
  \city{Guangzhou}
  \state{Guangdong}
  \country{China}}

\author{Dexi Liu}
\email{dexi.liu@163.com}
\orcid{0000-0003-1093-2744}
\author{Xiping Liu}
\email{liuxiping@jxufe.edu.cn}
\orcid{0000-0002-0230-8004}
\affiliation{%
  \institution{Jiangxi Key Laboratory of Data and Knowledge Engineering, China and School of Information Management, Jiangxi University of Finance and Economics}
  \city{Nanchang}
  \state{Jiangxi}
  \country{China}
}
\renewcommand{\shortauthors}{Wan, et al.}

\begin{abstract}

Document-level event extraction is a long-standing challenging information retrieval problem involving a sequence of sub-tasks: entity extraction, event type judgment, and event type-specific multi-event extraction. However, addressing the problem as multiple learning tasks leads to increased model complexity. Also, existing methods insufficiently utilize the correlation of entities crossing different events, resulting in limited event extraction performance. 
This paper introduces a novel framework for document-level event extraction, incorporating a new data structure called token-event-role and a multi-channel argument role prediction module. The proposed data structure enables our model to uncover the primary role of tokens in multiple events, facilitating a more comprehensive understanding of event relationships. By leveraging the multi-channel prediction module, we transform entity and multi-event extraction into a single task of predicting token-event pairs, thereby reducing the overall parameter size and enhancing model efficiency. 
The results demonstrate that our approach outperforms the state-of-the-art method by 9.5 percentage points in terms of the \textit{F}1 score, highlighting its superior performance in event extraction. Furthermore, an ablation study confirms the significant value of the proposed data structure in improving event extraction tasks, further validating its importance in enhancing the overall performance of the framework.

\end{abstract}



\maketitle

\section{Introduction}\label{sec:introduction}
\begin{figure*}
    \centering
    \includegraphics[width=\textwidth]{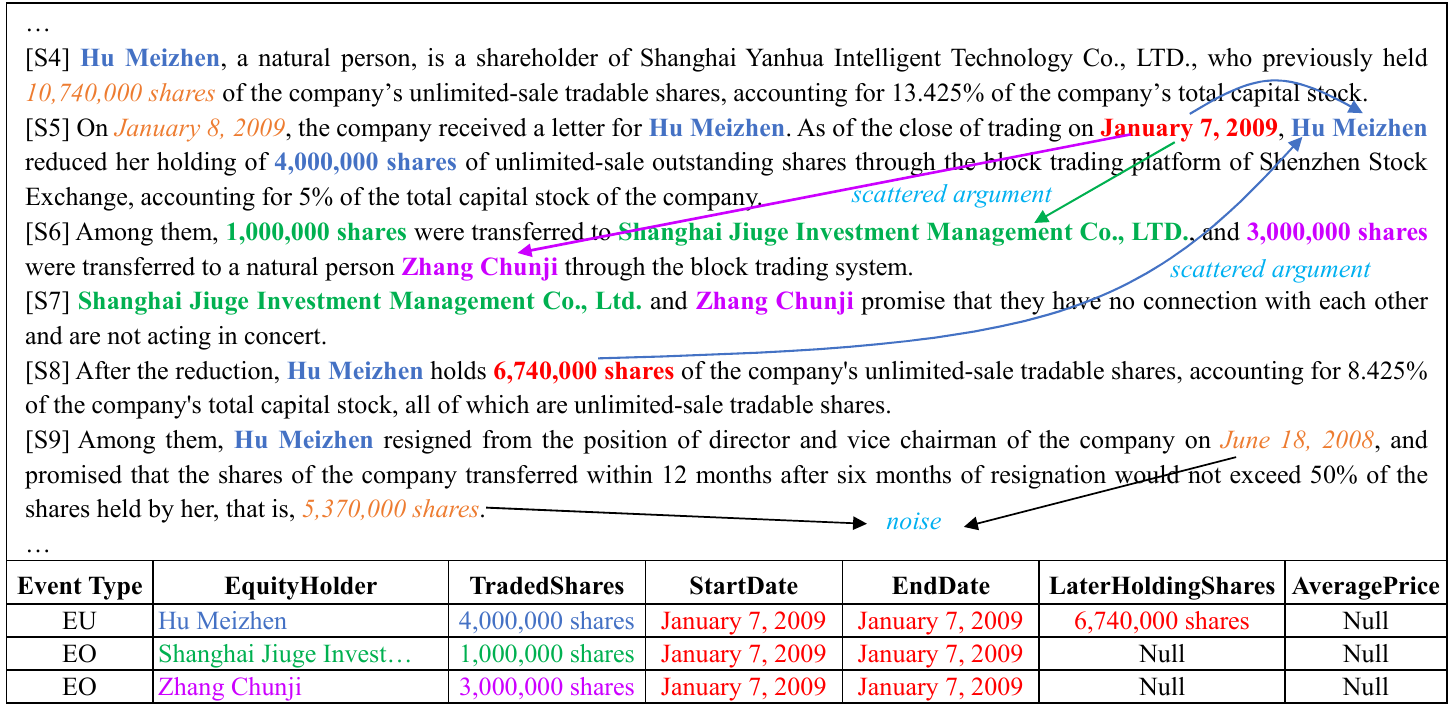}
    \caption{Document example containing three events with two event types. The top is the original document, and the bottom is the annotated event records without event triggers. EU and EO represent the event types of Equity Underweight and Equity Overweight, respectively. Bold tokens are event arguments. Red tokens refer to arguments shared by multiple events, and orange tokens represent the noise (not all marked).}
    \label{fig:example}
\end{figure*}

Document-level event extraction aims to identify events and extract corresponding arguments from a document that may include multiple events (See Fig. \ref{fig:example} as an example). 
Three important challenges must be addressed during the extraction process: multi-event detection, argument scattering, and multi-role argument identification.   
First, it is essential to accurately identify multiple distinct events at the document level.
Second, the task involves identifying arguments that span multiple sentences within an event.
Last, the model must handle scenarios where an entity plays various roles in both the same and different events. 
Since the linking triggers that connect arguments in events are often absent, this task necessitates a comprehensive understanding of the entire document, particularly the semantic correlation between arguments within an event or across different events.
The primary technical challenge is that the model must have an aggregated ability to assemble arguments across multiple sentences based on the semantic context of event arguments.

Previous studies on document-level event extraction \cite{Zheng2019,xu2021document,yang2021document,huang2021exploring,zhu2022efficient,liang2022raat} usually address the problem by splitting it into a sequence of sub-tasks: entity extraction, event type identification, and the recognition of multiple events and corresponding arguments based on identified event types (i.e., multi-event extraction). 

Given the multi-event extraction sub-task, Zheng et al. \cite{Zheng2019} developed a path extension strategy with entities as nodes to identify multiple events and corresponding arguments, forming an entity-based directed acyclic graph, in which each path corresponds to an event. Subsequently, Xu et al. \cite{xu2021document} and Liang et al. \cite{liang2022raat} explored other features based on this framework, respectively. Nevertheless, this method requires \textit{argument roles} (``roles'' for short) to be specified in advance for each event type, resulting in unsatisfactory extraction results and time-consuming training. To address this limitation, Yang et al. \cite{yang2021document} proposed a non-autoregressive parallel prediction network. Given an event type, each event record was modeled into a two-dimensional table of entity-role, and a binary classification was conducted to determine whether all entities in a document act as arguments in the current event based on the table. Due to the independent prediction for each event, this method separates the correlation information that an entity acts as arguments for the same/different roles in multiple events. 

To overcome the above limitations, this paper transforms the three sub-tasks into a single learning task, facilitated by a novel data structure called \textit{token-event-role}, and a multi-channel prediction module for role prediction. Specifically, by incorporating entity-event correspondences and matching relations between entity-event pairs and roles into the token-event-role data structure, we can directly capture the inherent relationships among entities (tokens), events, and roles. The new data structure not only facilitates the modeling of correlation semantics, where entities can act as arguments in different events, but also establishes a foundation for event extraction within a joint pattern. Additionally, we propose a multi-channel role prediction module to identify the role of each token-event pair given a specific event type (with each channel representing one event type). This approach eliminates the need for separate judgment sub-tasks, resulting in an efficient sing-task event extraction process.

To sum up, the main contributions of this work are threefold.

\begin{itemize}
    \item We introduce a novel data structure called \textit{token-event-role}, which efficiently captures the matching relations among tokens, events, and argument roles. This data structure directly reveals the roles played by tokens in events, providing a comprehensive representation of event information.
    \item We propose a multi-channel classification module for argument role  prediction for each token-event pair. By employing this module, we transform the previous three-step methods into a single-step multi-classification approach, resulting in a substantial reduction in parameter size and improved efficiency.
    \item Based on extensive experiments, our results confirm the effectiveness of our method in event extraction. Compared to state-of-the-art baselines, our model demonstrates significant improvements, achieving enhancements ranging from 9.5 to 33.2 percentage points in terms of the averaged \textit{F}1 score for event extraction. 
\end{itemize}

The remainder of this paper is structured as follows. In Section \ref{sec:data_structure}, we illustrate the data structure token-event-role, followed by a detailed illustration of our framework in Section \ref{sec:methodology}. After that, Section \ref{sec:experiments} demonstrates the experimental settings and overall results. In-depth analyses and discussions on our scheme are provided in Section \ref{sec:analyses}. In Section \ref{sec:related work}, we review the related work, and Section \ref{sec:conclusion} concludes the paper.  

\section{Token-event-role: A New Data Structure}\label{sec:data_structure}
This section mainly elaborates on the construction of the token-event-role and analyzes the performance of the data structure in theory. 

\subsection{Construction}
The construction of our token-event-role data structure mainly includes the following three steps. 

\textbf{Step 1: Numbering argument roles.} According to the event ontology, a number $r_k$ (i.e., tag) is assigned to the \textit{k}-th role contained in each event type. $0\leq k \leq |{\rm E}^t|_{\rm r}$, $|{\rm E}^t|_{\rm r}$ represents the number of argument roles in the \textit{t}-th event type ${\rm E}^t$. Role tag values are numbered from the integer 1 and increasing by 1. $r_0$=0 denotes the ``O'' tag, where ``O'' indicates the NULL in the role set of the current event type. 

\textbf{Step 2: Constructing token-event pairs.} Assume that the number of events contained in a corpus is \textit{m}, a hyper-parameter, and each event is represented by an event Id. Then, we combine each token with each event Id to form token-event pairs, denoted as $WE=\{(w_i\rule[2pt]{.3em}{.5pt}e_j)|1\leq i \leq n, 1\leq j \leq m\}$. \textit{n} indicates the number of tokens contained in a document, $w_i$ is the \textit{i}-th token, $e_j$ refers to the Id of the \textit{j}-th event.

\textbf{Step 3: Constructing the matching relations between token-event pairs and argument roles.} Given a token-event pair ($w_i\rule[2pt]{.3em}{.5pt}e_j$), the role of the token $w_i$ in the \textit{j}-th event $e_j$ is determined to be $r_k$ according to the annotated event record. The matching relations among all tokens, events, and argument roles under the \textit{t}-th event type $e^t$ based on the token-event-role data structure is $WER^t=\{(w_i\rule[2pt]{.3em}{.5pt}e_j\rule[2pt]{.3em}{.5pt}r_k)|1\leq i \leq n, 1\leq j \leq m, 0\leq k \leq |{\rm E}^t|_{\rm r}\}$, which serve as the gold tags for model training. 
If token $w_i$ does not act as an argument of any role in event $e_j$, the corresponding role tag of ($w_i\rule[2pt]{.3em}{.5pt}e_j$) pair is $r_0$=0. 

\begin{figure}
    \centering
    \includegraphics{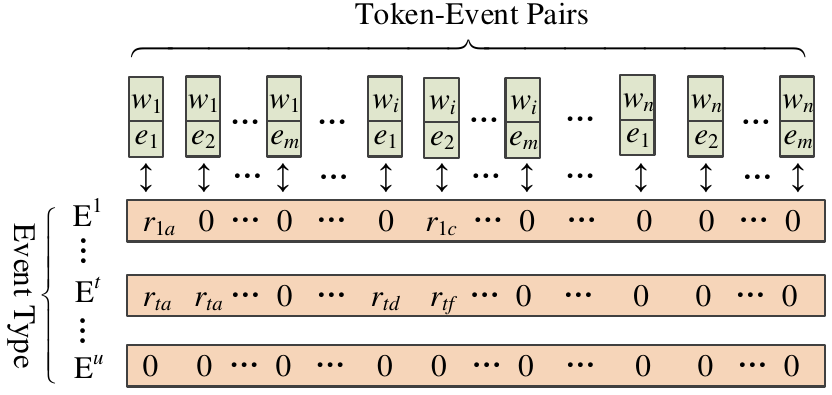}
    \caption{Token-event-role data structure construction. $r_{ta}$ is the \textit{a}-th role tag under the \textit{t}-th event type. $1\leq r_{1a}\leq |{\rm E^1}|_{\rm r}$, $1\leq r_{ta}\leq |{\rm E}^t|_{\rm r}$, $1\leq r_{ua}\leq |{\rm E}^u|_{\rm r}$. The number of roles under different event types may not be the same, i.e., $|{\rm E^1}|_{\rm r}$, $|{\rm E}^t|_{\rm r}$, and $|{\rm E}^u|_{\rm r}$ are not the same.}
    \label{fig:token-event-role}
\end{figure}

Regarding the multi-role argument issue, a token may act as an argument with diverse roles in the same event. It may also act as an argument with mixed ( the same or different) roles in multiple events. Thus, the data structure only distinguishes the situation of tokens in different events by the token-event correspondence. If token $w_i$ acts as arguments with different roles in the same event, multiple ($w_i\rule[2pt]{.3em}{.5pt}e_j$) pairs need to be constructed by duplicating token $w_i$, so that each ($w_i\rule[2pt]{.3em}{.5pt}e_j$) pair corresponds to a unique role tag $r_k$. 
Specifically, by observing the event ontology, the roles belonging to the above case should be determined first, and then the repeated tokens are identified by the technologies of token type recognition and semantic role labeling. Finally, according to the steps 2-3 above, the token-event-role matching relation set is constructed.

Fig. \ref{fig:token-event-role} shows the construction of the token-event-role matching relation. Given a document, according to the aforementioned steps, the matching relation \textit{WER} shown in Fig. \ref{fig:token-event-role} can be obtained. From the structure, it can be seen that each token is integrated with a specific event (green box), and the token's argument role in the event is clarified by the tag in the orange box. Thus, it can directly reveal the roles of tokens playing in events. 
Also, using the new data structure, the correlation semantics of an argument-playing token in events (e.g., which tokens belong to the same event, which roles of a token plays in different events) can be captured, and hence the learning and prediction ability should be improved.  

\begin{figure}
    \centering
    \includegraphics{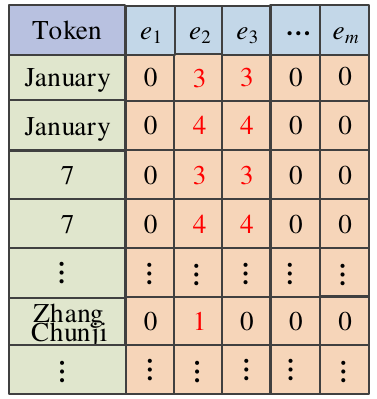}
    \caption{Matrix example of token-event pairs and corresponding role tag values for the event type EO in Fig. \ref{fig:example}. The argument roles under EO are showed in Fig. \ref{fig:example}, role tag values are numbered from left to right, event Ids are from top to bottom. The values of roles EquityHolder, StartDate, and EndDate are 1, 3, and 4, respectively. StartDate and EndDate roles belong to the multi-role for a token. As ``January'' is a time token, it is copied, and the two tokens correspond to the StartDate and EndDate role, respectively.}
    \label{fig:example-data-structure}
\end{figure}

To further illustrate the above construction mechanism, we demonstrate a matrix example for the EO event type in Fig. \ref{fig:example-data-structure}. As can be seen from Fig. \ref{fig:example}, the document contains three events, an EU-type event and two EO-type events. Given the EO event type, the event Ids are 2 and 3 according to the order of arrangement, corresponding to $e_2$ and $e_3$. Thus, 0 is filled in the other columns, including $e_1$ and $e_4$ to $e_m$. 
Since the token ``January'' plays both StartDate and EndDate roles in $e_2$, the role tag values 3 and 4 are filled in rows 1 and 2 of $e_2$; The situation is similar in $e_3$. 
As the token ``Zhang Chunji'' only plays the EquityHolder role in $e_2$; therefore the location of its row and $e_2$ column in the matrix is filled with tag value 1.

\subsection{Performance analysis}
In the following, we discuss the spatial cost of our scheme and existing document-level event extraction models in theory. Assume that the number of events contained in a corpus is \textit{m}, the number of argument roles for a given event type is \textit{r}, and a document contains \textit{n} entities. 

Given a document and an event type, DE-PPN \cite{yang2021document} generated an entity-role matrix for each event record; thus, its spatial complexity is $m \times n \times r$. Doc2EDAG \cite{Zheng2019} employed a path-expanding strategy, and the spatial complexity is $n \times p \times r$, where \textit{p} refers to the number of paths. Based on Doc2EDAG, GIT \cite{xu2021document} employed the graph convolutional network and added an adjacency matrix for the relations between entities, resulting in a $n \times p \times r$+$n \times n$ spatial complexity. ReDEE \cite{liang2022raat} was based on an ``\textit{entity+sentence}''-``\textit{entity+sentence}'' matrix, and the complexity is $(n+s) \times (n+s) \times r$, where \textit{s} represents the number of sentences in the document.

SCDEE \cite{huang2021exploring} and PTPCG \cite{zhu2022efficient} adopted the strategy of reducing candidate arguments. SCDEE first assigned the sentences in a document to \textit{m} communities (each community corresponds to an event), followed by argument role identification for each community. Therefore, the gold sentence-community matrix $s \times m$ and entity-role matrix $k \times r$ need to be given, and the final spatial complexity is $s \times m+m' \times k \times r$. Therein, $k \times r$ reveals the roles of \textit{k} entities in each event; $k \leq n$ represents the number of gold arguments in the event (it indicates the number of candidate arguments when training, which may be larger or smaller than the gold arguments); $m'$ denotes the number of gold events in the document and is different from \textit{m} (\textit{m} is a given hyper-parameter that reflects the number of events contained in the whole corpus). 
PTPCG used two matrices (i.e., entity-entity and entity-role) to identify multi-event and argument roles. Thus, the spatial complexity is $n \times n+m' \times k \times r$, where $n \times n$ is used to demonstrate whether there is an association between entities in the same event. 

For the data structure in this paper, if it is constructed for entities, the spatial complexity is $n \times m$, lower than existing models; if using tokens, corresponding complexity becomes $N \times m$, where \textit{N} is the number of tokens contained in the document. 
Considering those event arguments usually do not involve stop words, this paper filters them in the construction, reducing the spatial cost. Therefore, the spatial complexity of the data structure may be higher than that of previous methods through some pre-processing technologies, while the benefits are noticeable. The number of sub-tasks and model parameters is reduced. Also, the correlation semantics of tokens acting as arguments for the same/different roles in multiple events can be learned. 

\begin{figure*}
    \centering
    \includegraphics[width=\textwidth]{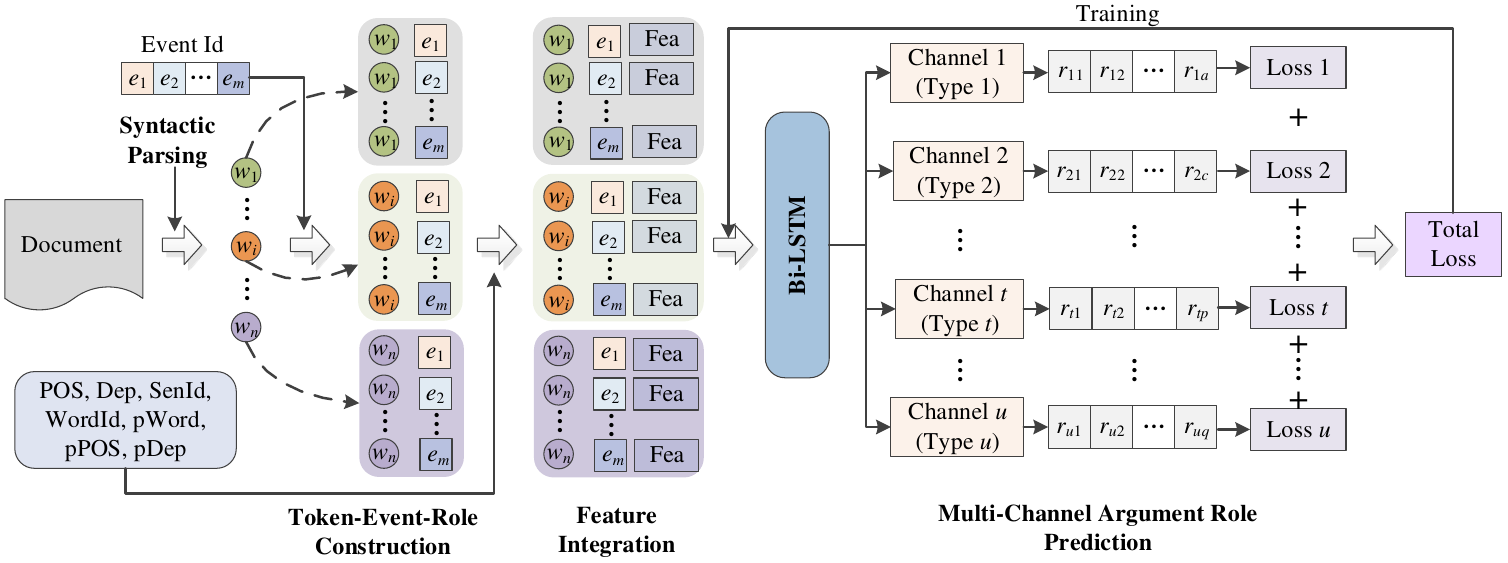}
    \caption{Token-Event-Role Structure-based Multi-Channel Event Extraction (TER-MCEE). Fea refers to other features, including POS, dependency relation (Dep), sentence location (SenId), token location (WordId), and the POS (pPOS) and Dep (pDep) of the parent.}
    \label{fig:frame}
\end{figure*}

\section{TER-MCEE: The Main Framework}\label{sec:methodology}
This section introduces our Token-Event-Role Structure-based Multi-Channel Event Extraction framework (TER-MCEE). According to the token-event-role data structure, the overall extraction task is integrated and transformed into a multi-classification problem of argument role types for predicting token-event pairs.

In the following, we describe our TER-MCEE framework. As illustrated in Fig. \ref{fig:frame}, the framework consists of five major components:

\begin{itemize}
\item \textit{Syntactic Parsing}: This component handles word segmentation, part-of-speech (POS) tagging, and dependency parsing.
\item \textit{Token-Event-Role Construction}: This component generates argument role tags for token-event pairs, leveraging the token-event-role data structure.
\item \textit{Feature Integration}: This component integrates various features associated with each token.
\item \textit{Bi-LSTM Layer}: The framework includes a Bi-LSTM layer that learns sequential semantics, capturing contextual information of the input sequence.
\item \textit{Multi-Channel Argument Role Prediction}: This component handles one event type in a channel and trains the model to predict argument role types of token-event pairs.
\end{itemize}

Together, these components form the foundation of the TER-MCEE framework, enabling effective event extraction based on the token-event-role structure.

\subsection{Syntactic Parsing}
Given a document, each sentence is successively sent to a syntactic parsing tool for word segmentation, POS tagging, and dependency parsing. Finally, the document is denoted as $d = \{w_1, \cdots, w_i, \cdots, \\w_n\}$, where \textit{n} represents the number of tokens in the document.  
In this paper, we use the pre-trained language model BERT \cite{Devlin2019} to initialize tokens' vectors, and the vector of the \textit{i}-th token $w_i$ is denoted as ${\rm \textbf{v}}_{{\rm Sem}_i}$.

\subsection{Token-Event-Role construction}
For the document \textit{d}, the matching relations between token-event pairs and argument roles are constructed according to the steps in Section 2; that is, argument role tags of token-event pairs are generated.

\subsection{Feature Integration}
To effectively identify event arguments, in addition to the sentence position feature \cite{Zheng2019,xu2021document}, we also consider the token position in sentences. Furthermore, the part-of-speech (POS) and dependency relation are essential clues, because arguments are mostly nouns and numerical tokens, while verbs are rarely served as; the dependency relations of arguments are mainly subject-verb or verb-object. Moreover, the parent information of the current node is meaningful in the dependency tree. 
As can be seen from Fig.\ref{fig:example}, not all numerical tokens in the document serve as arguments; thus, it is necessary to distinguish which numerical tokens serve as arguments and which do not. The major difference between them lies in the parent node they depend on.

\begin{figure}
    \centering
    \includegraphics{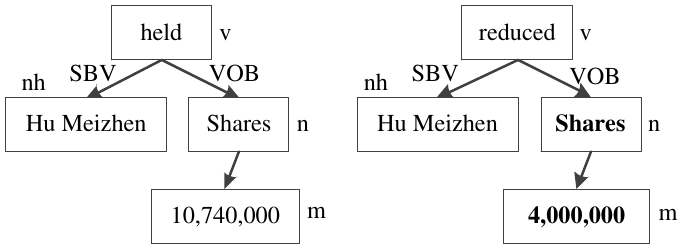}
    \caption{Partial dependency structures of sentences S4 and S5 in Fig. \ref{fig:example}. v, n, nh, and m refer to verb, general noun, person name, and number, respectively; SBV and VOB indicate the dependency relation types of subject-verb and verb-object.}
    \label{fig:dependency}
\end{figure}

Fig.\ref{fig:dependency} demonstrates partial dependency structures of sentences S4 and S5 in Fig. \ref{fig:example}. 
In Fig. \ref{fig:dependency}, the ``10,740,000'' and ``4,000,000'' subtrees guided by ``shares'' are of the same form, while the latter serves as an argument because its parent is ``reduced'' which triggers an event of Equity Underweight (EU) type, and the former's parent is ``held'' (No event triggered). 
Furthermore, the VOB dependency relation can reflect the role of node ``shares'' depending on the parent node ``reduced''. If it is not a VOB relation, the probability of node ``shares'' acting as an argument decreases.

Therefore, after integrating the above features, the final initial embedding of token-event pair ($w_i\rule[2pt]{.3em}{.5pt}e_j$) can be written as:
\begin{equation}
\begin{aligned}
    {\rm \textbf{v}}_{w_i\rule[2pt]{.3em}{.5pt}e_j} =\; &{\rm \textbf{v}}_{{\rm Sem}_i}\;||\;{\rm \textbf{v}}_{e_j}\;||\;{\rm \textbf{v}}_{{\rm POS}_i}\;||\;{\rm \textbf{v}}_{{\rm Dep}_i}\;||\;{\rm \textbf{v}}_{{\rm SenId}_i}\\
    &||\;{\rm \textbf{v}}_{{\rm WordId}_i}\;||\;{\rm \textbf{v}}_{{\rm pWord}_i}\;||\;{\rm \textbf{v}}_{{\rm pPOS}_i}\;||\;{\rm \textbf{v}}_{{\rm pDep}_i}
\end{aligned},
\end{equation}
where SenId indicates the sentence Id of a token $w_i$ residing, and WordId refers to the position Id of $w_i$. pWord, pPOS, and pDep represent the token, POS, and dependency relation of the parent node, respectively. 
In addition to ${\rm \textbf{v}}_{{\rm Sem}_i}$, other vectors are generated by looking up the randomly initialized embedding table. $\parallel$ is the concatenation operation.

\subsection{Bi-LSTM}
For each token-event pair ($w_i\rule[2pt]{.3em}{.5pt}e_j$), a forward LSTM encodes ($w_i\rule[2pt]{.3em}{.5pt}e_j$) by learning the contextual information from ($w_1\rule[2pt]{.3em}{.5pt}e_1$) to ($w_i\rule[2pt]{.3em}{.5pt}e_j$), marked as $\overrightarrow{{\rm \textbf{h}}}_{w_i\rule[2pt]{.3em}{.5pt}e_j}$, followed by a backward LSTM encoding ($w_i\rule[2pt]{.3em}{.5pt}e_j$) based on the contextual information from ($w_n\rule[2pt]{.3em}{.5pt}e_m$) to ($w_i\rule[2pt]{.3em}{.5pt}e_j$), which is marked as $\overleftarrow{{\rm \textbf{h}}}_{w_i\rule[2pt]{.3em}{.5pt}e_j}$. 
The details are shown as follows:
\begin{equation}\label{eq2}
    \overrightarrow{{\rm \textbf{h}}}_{w_i\rule[2pt]{.3em}{.5pt}e_j} = \overrightarrow{\rm LSTM}\left({\rm \textbf{v}}_{w_1\rule[2pt]{.3em}{.5pt}e_1}, {\rm \textbf{v}}_{w_1\rule[2pt]{.3em}{.5pt}e_2}, \cdots, {\rm \textbf{v}}_{w_i\rule[2pt]{.3em}{.5pt}e_j}\right),
\end{equation}
\begin{equation}\label{eq3}
    \overleftarrow{{\rm \textbf{h}}}_{w_i\rule[2pt]{.3em}{.5pt}e_j} = \overleftarrow{\rm LSTM}\left({\rm \textbf{v}}_{w_i\rule[2pt]{.3em}{.5pt}e_j}, {\rm \textbf{v}}_{w_i\rule[2pt]{.3em}{.5pt}e_{j+1}}, \cdots, {\rm \textbf{v}}_{w_n\rule[2pt]{.3em}{.5pt}e_m}\right).
\end{equation}
Finally, $\overrightarrow{{\rm \textbf{h}}}_{w_i\rule[2pt]{.3em}{.5pt}e_j}$ and $\overleftarrow{{\rm \textbf{h}}}_{w_i\rule[2pt]{.3em}{.5pt}e_j}$ are concatenated to represent the embedding of token-event pair ($w_i\rule[2pt]{.3em}{.5pt}e_j$) after Bi-LSTM, denoted as ${\rm \textbf{h}}_{w_i\rule[2pt]{.3em}{.5pt}e_j} = \overrightarrow{{\rm \textbf{h}}}_{w_i\rule[2pt]{.3em}{.5pt}e_j} \parallel \overleftarrow{{\rm \textbf{h}}}_{w_i\rule[2pt]{.3em}{.5pt}e_j}$.

\subsection{Multi-Channel Argument Role Prediction}
To distinguish different event types, we exploit a multi-channel mechanism to make each channel correspond to an event type. The initial inputs of all channels are ${\rm \textbf{h}}_{w_i\rule[2pt]{.3em}{.5pt}e_j}$ output by of Bi-LSTM layer. In each channel, a fully-connected neural network (FCNN) is used to form the embedding representation of the corresponding dimension according to the number of argument roles under the current channel. Mathematically, the embedding representation of ${\rm \textbf{h}}_{w_i\rule[2pt]{.3em}{.5pt}e_j}$ output by FCNN for the \textit{t}-th channel can be formulated as: 
\begin{equation}\label{eq4}
    {\rm \textbf{h}}_{w_i\rule[2pt]{.3em}{.5pt}e_j}^{t}={\rm FCNN}\left({\rm \textbf{h}}_{w_i\rule[2pt]{.3em}{.5pt}e_j}||\;{\rm \textbf{h}}_{t}\right),
\end{equation}
where ${\rm \textbf{h}}_{t}$ denotes the embedding representation of the \textit{t}-th event type whose initial vector is also generated by looking up the randomly initialized embedding table.

After the multi-channel learning, each token-event pair obtains the embedding representation ${\rm \textbf{h}}_{w_i\rule[2pt]{.3em}{.5pt}e_j}^{t}$ for different channels, then we employ a softmax function to compute the distribution $p(y_{w_i\rule[2pt]{.3em}{.5pt}e_j}|\theta)$ over argument role tags under current channel:
\begin{equation}\label{eq5}
    p\left(y_{w_i\rule[2pt]{.3em}{.5pt}e_j}|\theta \right) = {\rm softmax}\left({\rm \textbf{W}}_p\, {\rm \textbf{h}}_{w_i\rule[2pt]{.3em}{.5pt}e_j}^{t} + b_p\right),
\end{equation}
where $y_{w_i\rule[2pt]{.3em}{.5pt}e_j}$ is the argument role tag of token-event pair ($w_i\rule[2pt]{.3em}{.5pt}e_j$) under the parameter $\theta$, ${\rm \textbf{W}}_p$ denotes a weight matrix, mapping ${\rm \textbf{h}}_{w_i\rule[2pt]{.3em}{.5pt}e_j}^{t}$ to the feature score for each class label, and $b_p$ is a bias term. After softmax, the class tag with the largest probability is chosen as the classification result.

Given that the number of ``O'' role tags is much larger than that of gold argument role tags, following previous work \cite{Chen2018,Cui2020,wan2022multi}, the standard cross-entropy loss with weight is used as our objective function to strengthen the influence of gold role tags:
\begin{equation}\label{eq6}
    J\left(\theta \right) = -\sum_{i=1}^{n}\sum_{j=1}^{m}\omega_{w_i\rule[2pt]{.3em}{.5pt}e_j} {\rm log}\,p\left(y_{w_i\rule[2pt]{.3em}{.5pt}e_j}|\theta \right),
\end{equation}
where \textit{n} represents the number of tokens in the document, \textit{m} is the number of events contained in a corpus, $\omega_{w_i\rule[2pt]{.3em}{.5pt}e_j}$ is the weight of $y_{w_i\rule[2pt]{.3em}{.5pt}e_j}$ tag, which can be computed according to the Equation (\ref{eq7}) in \cite{wan2022multi}. 
\begin{equation}\label{eq7}
    \omega_{w_i\rule[2pt]{.3em}{.5pt}e_j} = \frac{{\rm Med}\left(N_{C_1}, N_{C_2}, \cdots, N_{C_n}\right)}{N_{C_r}},
\end{equation}
where Med($\cdot$) means taking the median operation, $N_{C_r}$ is the number of tokens contained in the \textit{r}-th class in a corpus, and $C_r$ is the \textit{r}-th class tag. 

Finally, we add the loss of each channel to obtain the total loss of the model, which is used to guide model training:
\begin{equation}
    L_{\rm Total}\;=\;L_{1}\;+\;L_{2}\;+\;\cdots\;+\;L_{u},
\end{equation}
where \textit{u} is the number of event types in the corpus.

\section{Experiments and Results}\label{sec:experiments}
\subsection{Dataset and Evaluation Metrics}
This paper conducted experiments on a widely used corpus ChFinAnn \cite{Zheng2019}. It consists of financial announcements, a total of 32,040 documents covering five event types: Equity Freeze (EF), Equity Repurchase (ER), Equity Underweight (EU), Equity Overweight (EO), and Equity Pledge (EP).  
We followed the standard split of the dataset, 25,632/3,204/3,204 documents for training/development/test set. A document in the dataset contains 20 sentences on average, and 29\% of documents involve multiple events. 
To verify the robustness of our scheme on a small dataset, we randomly sampled the dataset, and the number of documents for training/development/test is 200/100/100, respectively. The statistical results are reported in Table \ref{tab:sample dataset}.

\begin{table}
    \centering
    \caption{Sample Dataset Statistics about the Number of Documents for the Train, Development (Dev) and Test}
    \label{tab:sample dataset}
    \begin{tabular}{ccccc}
    \toprule
        \textbf{Event} & \textbf{Train} & \textbf{Dev} & \textbf{Test} & \textbf{Total} \\
    \midrule     
        EF & 119 & 57 & 66 & 242 \\
        ER & 149 & 84 & 93 & 326 \\
        EU & 168 & 80 & 76 & 324 \\
        EO & 134 & 74 & 75 & 283 \\
        EP & 137 & 61 & 62 & 260 \\
    \midrule
        Total & 707 & 356 & 372 & 1,435 \\
    \bottomrule
    \end{tabular}
\end{table}

We used the LTP\footnote{http://ltp.ai/index.html} syntactic tool to parse the sentences in documents. 
Regarding experimental evaluation metrics, the Precision (\textbf{\textit{P}}), Recall (\textbf{\textit{R}}) and \textit{F}1-score (\textbf{\textit{F}1}) were adopted to evaluate the models.
Since a document contains enormous tokens, and few of them serve as gold argument roles, the model performed well for those not-arguments (i.e., the tokens marked as ``O''). However, calculating the overall value of \textit{F}1 score can not accurately reflect the recognition effect of arguments. Therefore, the prediction results of the ``O'' tag were filtered out in the evaluation.

\subsection{Hyper-parameters and Baselines}
We choose the Adam optimizer in experiments, and set batch size = 1, learning rate = 1e-3, dropout = 0.2, and iteration = 10. The embedding dimensions of tokens and other features are 768 and 50. The hidden layer size and layers of Bi-LSTM is 200 and 4. The number of events contained in coupus is set to 34, which is the maximum number of events. 
For the small sample dataset, the iteration is set to 100. 
Experimental environment of this paper is Python3.7, PyTorch1.10, and NVIDIA GeForce RTX 3090 24G. 

To comprehensively evaluate our TER-MCEE, we compare it with a range of state-of-the-art models. 
\textbf{DCFEE} \cite{Yang2018} developed a key-event sentence detection to extract arguments from key-event mention and surrounding sentences. The model has two variants: DCFEE-O only produced one event record from one key-event sentence, and DCFEE-M extracted multiple events from a document. 
\textbf{Doc2EDAG} \cite{Zheng2019} transformed the document-level event extraction as directly filling event tables with entity-based path expanding. \textbf{Greedy-Dec} is a simple baseline of Doc2EDAG, which only fills one event table entry greedily. 
\textbf{GIT} \cite{xu2021document} designed a heterogeneous graph-based interaction model with a tracker to capture global interactions among various sentences and entity mentions. 
\textbf{DE-PPN} \cite{yang2021document} proposed a document-level encoder and a multi-granularity decoder to extract all events in parallel. 
\textbf{SCDEE} \cite{huang2021exploring} introduced sentence community and assigned all sentences of an event to a sentence community. 
\textbf{PTPCG} \cite{zhu2022efficient} constructed the maximal clique by calculating pseudo-trigger and incorporated other common entities to complete the clique, ensuring all entities of an event are in one clique. 
\textbf{ReDEE} \cite{liang2022raat} integrated the correlations between event arguments in an event and designed an argument relation attention encoder.

\subsection{Overall Performance}
The overall performance results are reported in Table \ref{tab:overall result}. ``Params'' represents the number of parameters (1M=10$^6$) contained in the model, which is used to measure the size of the model. $*$ indicates the reproduced results using corresponding open-source code; $\dag$ refers to the results from Zhu et al. \cite{zhu2022efficient}; $\ddag$ indicates results from Zheng et al. \cite{Zheng2019}; Other results are from the original literature. Since the GPU capacity in our experimental environment does not support ReDEE implementation, the batch size of its training set is modified to 16. Avg represents the average result of all event types.  

\begin{table*}
    \centering
    \caption{Overall \textit{F}1 Score Evaluated on the Test Set}
    \label{tab:overall result}
    \begin{tabular}{cccccccc}
    \toprule
        \textbf{Model} & \textbf{Params(M)} & \textbf{EF} & \textbf{ER} & \textbf{EU} & \textbf{EO} & \textbf{EP} & \textbf{Avg} \\
    \midrule     
        DCFEE-O$^\ddag$ & 32$^\dag$ & 51.1 & 83.1 & 45.3 & 46.6 & 63.9 & 58.0 \\
        DCFEE-M$^\ddag$ & 32$^\dag$ & 45.6 & 80.8 & 44.2 & 44.9 & 62.9 & 55.7 \\
        Greedy-Dec$^\ddag$ & 64$^\dag$ & 58.9 & 78.9 & 51.2 & 51.3 & 62.1 & 60.5 \\
        Doc2EDAG & 64$^\dag$ & 70.2 & 87.3 & 71.8 & 75.0 & 77.3 & 76.3 \\
        GIT & 97$^\dag$ & 73.4 & 90.8 & 74.3 & 76.3 & 77.7 & 78.5 \\
        DE-PPN & 130.38$^*$ & 73.5 & 87.4 & 74.4 & 75.8 & 78.4 & 77.9 \\
        SCDEE & 93.45$^*$ & 80.4 & 90.5 & 75.1 & 70.1 & 78.1 & 78.8 \\
        PTPCG & 32$^\dag$ & 71.4 & 91.6 & 71.5 & 72.2 & 76.4 & 76.6 \\
        ReDEE & 201.8$^*$ & 74.1 & 90.7 & 75.3 & 78.1 & 80.1 & 79.4 \\
        \textbf{TER-MCEE} & \textbf{5.26} & \textbf{87.9} & \textbf{97.2} & \textbf{89.8} & \textbf{91.2} & \textbf{78.6} & \textbf{88.9} \\
    \bottomrule
    \end{tabular}
\end{table*}

As shown in Table \ref{tab:overall result}, our TER-MCEE consistently outperforms other baselines, with the Avg increased by 9.5$\sim$33.2 percentage points. This is because all baselines used the pipeline pattern for the whole extraction task, resulting in the error propagation. Also, pipelined implementation makes it difficult for the model to approach the training goal, e.g., the correlation semantics of a token acts as arguments in different events can not be captured due to separating multiple subtasks, reducing baselines' extraction effect. The detailed analysis is as follows.

\textbf{Error propagation for entity extraction}. By analyzing the results of baselines in each phase, it is found that there are many errors in entity extraction \cite{zhu2022efficient}. Especially for financial data that contains abundant numerical words (e.g., money, data, percentage ratio, and shares), a common model for entity extraction is challenging in identifying such entities. 
Table \ref{tab:entity extraction} reports the entity extraction effect of baselines, and the results are obtained from the supplementary material of literature \cite{xu2021document}. \textit{P} and \textit{R} represent precision and recall, respectively. 
In Table \ref{tab:entity extraction}, there are more than 10 percentage points of errors in baselines, directly affecting the subsequent argument role identification. When the dataset is small, the training samples are insufficient to support the learning of the model in all phases, resulting in unfavorable results.

\textbf{Error propagation of entity-event correspondence}. For SCDEE and PTPCG, in addition to entity extraction and event type judgement errors, there are also errors in the assignment of sentence to community and event decoding by the clique, respectively. 
According to the supplement material provided by Zhu et al. \cite{zhu2022efficient}, 14.6 percentage points of errors have been found in the entity-entity gold matrix of PTPCG when decoding events. Thus, there may be more errors based on the prediction matrix output by the model. In the experiment, the precision, recall, and \textit{F}1 score of argument combination are all only 40.9\%. This indicates that many entities acting as the arguments of an event are not in the same clique, and each clique includes numerous entities of other cliques; that is, the construction of entity-event correspondence is not accurate enough. These factors are also the primary reasons why its Avg is inferior to that of GIT and SCDEE. 

\begin{table}
    \centering
    \caption{\textit{F}1 Score of Baselines for Entity Extraction}
    \label{tab:entity extraction}
    \begin{tabular}{cccc}
    \toprule
        \textbf{Model} & \textbf{\textit{P}} & \textbf{\textit{R}} & \textbf{\textit{F}1} \\
    \midrule
        DCFEE-O & 86.5 & 88.6 & 87.6 \\
        DCFEE-M & 86.6 & 89.0 & 87.8 \\
        Greedy-Dec & 87.5 & 89.8 & 88.6 \\
        Doc2EDAG & 88.0 & 90.0 & 890 \\
        GIT & 85.8 & 92.6 & 89.1 \\
    \bottomrule
    \end{tabular}
\end{table}

\textbf{Intermediate phases of baselines}. Due to similar intermediate phases, baselines' performances on Avg are comparative. GIT captured and encoded the structure information in the document-level heterogeneous interaction graph; thus, it outperformed Doc2EDAG with 2.2 percentage points on Avg. 
SCDEE benefits from the divided sentence community. Compared with other baselines that treat all entities in a document as candidate arguments, SCDEE reduced the number of candidate arguments, reducing the difficulty of training the model to determine whether an entity acts as an argument in a given event. Therefore, it achieves the second effect in baselines. Since the correlation information between entities is a key clue to assemble entities into events, and ReDEE encoded the information based on the framework of Doc2EDAG, it beats other baselines.
Nevertheless, all baselines are pipeline patterns, and the error propagation restricts their performances by only about 77\%, which is much lower than our joint model. 

\textbf{Additional features}. Sentence position and entity type are used in most baselines. Regarding these two features, the former is retained herein. Nevertheless, this feature does not play a fundamental role in the entire extraction according to the ablation experiment in Table \ref{tab:ablation}. 
Consistent with human understanding, it is challenging to distinguish the role of an entity playing according to the sentence position. 
Compared with sentence position, part-of-speech (POS), parent of dependency, and dependency relation are more beneficial to identifying which entities serve as arguments. Fig. \ref{fig:dependency} explicitly reveals the impact of these features from a theoretical perspective.

Regarding model parameters, the corresponding results are provided in Table \ref{tab:overall result}, further supporting the theoretical analysis mentioned earlier. TER-MCEE exhibits a significantly lower parameter count compared to baselines that involve at least three sub-tasks. With only 5.26M parameters, TER-MCEE outperforms the baselines that have larger parameter sizes.

In summary, the strong performance of our TER-MCEE framework can be attributed to the following three factors.
First, the \textit{token-event-role} data structure plays a crucial role in our framework. It effectively reveals the roles played by tokens in events and enables the transformation of sub-tasks, including entity extraction and multi-event extraction, based on recognized event types, into a multi-classification problem for predicting argument role types in token-event pairs. Moreover, this data structure leverages the correlations among tokens within the same event and across different events, which enhances the effectiveness of document-level event extraction.
Second, our TER-MCEE framework incorporates the \textit{multi-channel argument role prediction} module, which integrates multiple sub-tasks into a single prediction task in each channel. This design reduces the number of model parameters required while maintaining performance.
Last, TER-MCEE utilizes key features that contribute to its performance. For example, the incorporation of a parent token, combined with part-of-speech (POS) and dependency relations, facilitates event-type recognition.

\section{Additional Analysis and Discussions}\label{sec:analyses}
To further investigate the specific impact of each component on event extraction performance, we conduct additional experiments, including single-event and multi-event, ablation, and small sample. 

\subsection{Single-Event and  Multi-Event}
This section aims to analyze the extraction effects of models when a document contains only one or more events. Table \ref{tab:single multiple} reports the \textit{F}1 score of each model under single-event (S.) and multi-event (M.). $\dag$ refers to results from Xu et al. \cite{xu2021document}, and the other results are taken from the original paper.

\begin{table*}[t]
    \centering
    \caption{\textit{F}1 Score of Single-Event and Multi-Event}
    \label{tab:single multiple}
    \resizebox{.9\linewidth}{!}{
    \begin{tabular}{ccccccccccccc}
    \toprule
        \multirow{2}{*}{\textbf{Model}} & \multicolumn{2}{c}{\textbf{EF}} & \multicolumn{2}{c}{\textbf{ER}} & \multicolumn{2}{c}{\textbf{EU}} & \multicolumn{2}{c}{\textbf{EO}} & \multicolumn{2}{c}{\textbf{EP}} & \multicolumn{2}{c}{\textbf{Avg}} \\ \cmidrule(lr){2-3} \cmidrule(lr){4-5} \cmidrule(lr){6-7} \cmidrule(lr){8-9} \cmidrule(lr){10-11} \cmidrule(lr){12-13}
        {} & \textbf{S.} & \textbf{M.} & \textbf{S.} & \textbf{M.} & \textbf{S.} & \textbf{M.} & \textbf{S.} & \textbf{M.} & \textbf{S.} & \textbf{M.} & \textbf{S.} & \textbf{M.} \\
    \midrule     
        DCFEE-O$^\dag$ & 55.7 & 38.1 & 83.0 & 55.5 & 52.3 & 41.4 & 49.2 & 43.6 & 62.4 & 52.2 & 60.5 & 46.2 \\
        DCFEE-M$^\dag$ & 45.3 & 40.5 & 76.1 & 50.6 & 48.3 & 43.1 & 45.7 & 43.3 & 58.1 & 51.2 & 54.7 & 45.7 \\
        Greedy-Dec$^\dag$ & 74.0 & 40.7 & 82.2 & 50.0 & 61.5 & 35.6 & 63.4 & 29.4 & 78.6 & 36.5 & 71.9 & 38.4 \\
        Doc2EDAG$^\dag$ & 79.7 & 63.3 & 90.4 & 70.7 & 74.7 & 63.3 & 76.1 & 70.2 & 84.3 & 69.3 & 81.0 & 67.4 \\
        GIT & 81.9 & 65.9 & 93.0 & 71.7 & 82.0 & 64.1 & 80.9 & 70.6 & 85.0 & 73.5 & 84.6 & 69.2 \\
        DE-PPN & 82.1 & 63.5 & 89.1 & 70.5 & 79.7 & 66.7 & 80.6 & 69.6 & \textbf{88.0} & 73.2 & 83.9 & 68.7 \\
        PTPCG & 83.6 & 59.9 & 93.7 & 73.8 & 77.3 & 63.6 & 79.7 & 62.8 & 86.1 & 70.5 & 84.1 & 66.1 \\
        ReDEE & 79.1 & 69.1 & 92.7 & 73.6 & 79.9 & 69.2 & 81.6 & 73.7 & 86.3 & \textbf{76.5} & 83.9 & 72.4 \\
        \textbf{TER-MCEE} & \textbf{89.9} & \textbf{85.6} & \textbf{97.8} & \textbf{91.5} & \textbf{93.7} & \textbf{87.3} & \textbf{92.3} & \textbf{90.5} & 86.9 & 75.2 & \textbf{92.1} & \textbf{86.0} \\
    \bottomrule
    \end{tabular}}
\end{table*}

In Table \ref{tab:single multiple}, for single-event and multi-event, our model (TER-MCEE) outperforms all baselines in most event types and Avg, demonstrating that TER-MCEE is as effective in dealing with a single event and multi-event separately. Concretely, the Avg on single-event is superior to baselines with 7.5$\sim$37.4 percentage points, and a 13.6$\sim$47.6 percentage points increase is achieved on the Avg of multi-event. 
Compared with Table \ref{tab:overall result}, it can be seen that the overall effect trend is consistent across baselines. GIT and DE-PPN perform well, with a slight distinction. 

For the EP event type, the statistics of experimental datasets show that the multiple events related to this type are basically EP type, accounting for 99.84\% (i.e., only this type is included in the document). Thus, the extraction difficulty decreases more, and the differences between the models are small.
As DE-PPN does not need to specify argument role order, it has the highest effect on single-event and is comparable to GIT on multi-event. 
Looking at the Avg of Greedy-Dec, it was found that the model could identify single-event well but was not good at extracting multiple events, given that the \textit{F}1 score on multi-event was only 38.4\%.

\subsection{Ablation}
In this paper, many features are used and encoded. To verify the validity of these features, we conducted feature ablation experiments, as shown in Table \ref{tab:ablation}. POS and Dep refer to the part-of-speech and dependency relation of current token, and pPOS and pDep are corresponding meanings for the parent node. SenId denotes the position of the sentence containing the current token, and WordId is the position of current token. eId and eType represent the number of event and event type. ``w/o All'' means that all additional features are moved and only the semantic embedding representation of current token is retained. 

\begin{table}
    \centering
    \caption{\textit{F}1 Score of Feature Ablation}
    \label{tab:ablation}
    \begin{tabular}{cccccccc}
    \toprule
        \textbf{No.} & \textbf{Ablation} & \textbf{EF} & \textbf{ER} & \textbf{EU} & \textbf{EO} & \textbf{EP} & \textbf{Avg} \\
    \midrule     
        1 & Ours & 87.9 & 97.2 & 89.8 & 91.2 & 78.6 & 88.9 \\
        2 & w/o POS & 68.0 & 96.6 & 87.6 & 88.7 & 80.6 & 84.3 \\
        3 & w/o Dep & 82.2 & 96.7 & 87.6 & 86.5 & 80.2 & 86.6 \\
        4 & w/o SenId & 81.9 & 97.1 & 89.1 & 90.4 & 80.6 & 87.8 \\
        5 & w/o WordId & 82.2 & 97.7 & 85.6 & 90.1 & 82.2 & 87.6 \\
        6 & w/o pWord & 79.9 & 97.3 & 87.3 & 90.6 & 82.3 & 87.5 \\
        7 & w/o pPOS & 78.4 & 96.8 & 84.1 & 90.0 & 80.5 & 86.0 \\
        8 & w/o pDep & 78.6 & 96.1 & 83.7 & 87.2 & 78.7 & 84.9 \\
        9 & w/o eId & 77.0 & 96.5 & 80.6 & 86.5 & 78.9 & 83.9 \\
        10 & w/o eType & 80.2 & 95.4 & 86.2 & 82.9 & 80.3 & 85.0 \\
        11 & w/o All & 76.8 & 91.2 & 84.3 & 83.2 & 79.1 & 82.9 \\
    \bottomrule
    \end{tabular}
\end{table}

As Table \ref{tab:ablation} shows, all features proposed are valuable for event extraction, and removing them reduces the Avg by 1.1$\sim$6.0 percentage points. Compared with Avg in lines 2 and 3, it can be seen that POS plays a greater role in the extraction, and the Avg decreases by 2.3 percentage points more than Dep. This is consistent with the analysis above because arguments are mainly nouns and numerals, which are easier to identify by POS. 
However, the corresponding results of the parent on pPOS (line 7) and pDep (line 8) are reversed. When removing pDep, its effect is lower than that of removing pPOS by 1.1 percentage points. This is because the parent node needs to depend on the dependency relation to decide whether it acts as a predicate. Only acting as a predicate, an event may be triggered, and tokens depending on the predicate can be used as event arguments. 

Observing line 9, it is found that the event Id (eId) plays a key role. If the eId is not considered, Avg is only 83.9\%, with a decrease of 5.0 percentage points. 
This is because the data structure proposed integrates the token-event correspondences and the matching relations between token-event pairs and roles, effectively capturing the correlation semantics of an token acting as arguments in different events. 
However, Yang et al. \cite{yang2021document} separated this information. This is the main reason why it is inferior to our model. 
Compared with eId, eliminating eType (line 10) results in a smaller reduction in efficacy than eId. Nevertheless, given that the Avg is only 85.0\%, eType is an essential clue in the document-level event extraction because it could explicitly distinguish different event types.

\subsection{Small Sample}
\begin{figure}
    \centering
    \includegraphics{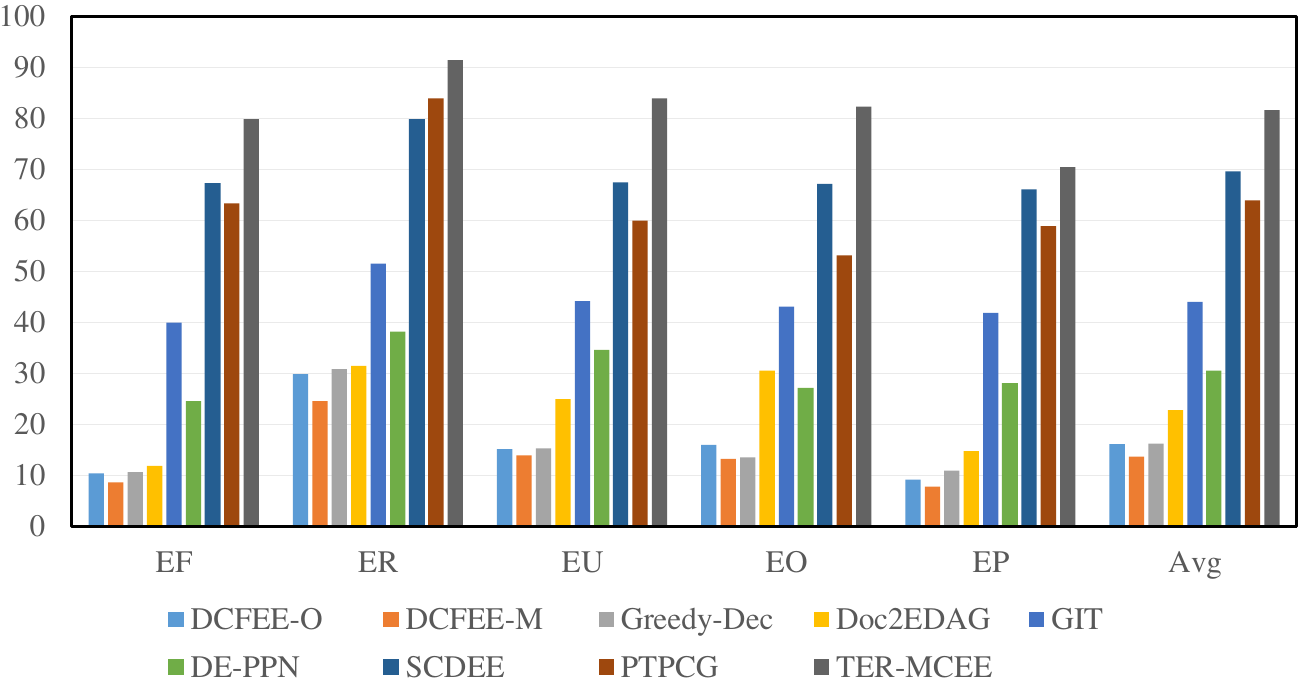}
    \caption{\textit{F}1 score of models on the small sample dataset.}
    \label{fig:small sample}
\end{figure}

Since the model in this paper avoids a serial prediction, such as entity extraction, event type judgment, and sentence community detection, it may perform well on small samples. Therefore, a small sample experiment was conducted to confirm the robustness of the model. Fig. \ref{fig:small sample} demonstrates the \textit{F}1 score of each baseline and TER-MCEE on a small sample dataset. 

As illustrated in Fig. \ref{fig:small sample}, TER-MCEE remains excellent performance on the small sample, with the Avg reaching 81.6\%, 12.0$\sim$68.0 percentage points above baselines. These results suggest the robustness of our model and reflect that baselines' performance highly relies on the training samples, hence significantly restricting the applicability of baselines. 
As expected, the token-event-role data structure contributes to outstanding results because the model can directly perform the training according to the ultimate goal matrix using the data structure, without adjusting model parameters in each phase. Therefore, the training target can be approached quickly by regulating a few parameters on a small sample dataset. 
 
Observing the baselines in Fig. \ref{fig:small sample}, the models that adopt the path generation have an undesired performance when the sample is small. 
For autoregressive models such as Doc2EDAG and GIT, a decreasing training sample scale tends to result in worse performance in generating paths on a specified role order. 
Table \ref{tab:argument role on sample} reports the identification quality of each role of the Doc2EDAG model under two event types (EU and EO) on the small sample dataset. 
Based on the role order in Table \ref{tab:argument role on sample}, the \textit{F}1 score of argument identification on ``EquityHolder'' is over 63.9\%, and the performance decreases for subsequent argument roles. This indicates that the order-based strategy is sensitive to the scale of the training dataset, and favorable learning results can be achieved only on a large-scale dataset. 
Although DE-PPN does not need the specific order, the scope of candidate entities is still a full document in the argument identification, leading to an unsatisfactory performance on the small sample dataset. 

\begin{table}
    \centering
    \caption{\textit{F}1 Score of Each Argument Role under EU and EO Event Type}
    \label{tab:argument role on sample}
    \begin{tabular}{ccccccc}
    \toprule
        \multirow{2}{*}{\textbf{Role Type}} & \multicolumn{3}{c}{\textbf{EU}} & \multicolumn{3}{c}{\textbf{EO}} \\ 
        \cmidrule(lr){2-4} \cmidrule(lr){5-7}
        {} & \textbf{\textit{P}} & \textbf{\textit{R}} & \textbf{\textit{F}1} & \textbf{\textit{P}} & \textbf{\textit{R}} & \textbf{\textit{F}1} \\
    \midrule     
        EquityHolder & 60.9	& 76.4 & 67.8 & 68.1 & 60.3 & 63.9 \\
        TradedShares & 51.5 & 45.2 & 48.1 & 66.2 & 19.7 & 30.4 \\
        StartDate & 55.3 & 44.9 & 49.6 & 66.9 & 50.6 & 57.6 \\
        EndDate & 55.7 & 46.3 & 50.6 & 64.2 & 46.1 & 53.7 \\
        LaterHoldingShares & 24.8 & 28.2 & 26.4 & 49.2 & 8.5 & 14.5 \\
        AveragePrice & 58.3 & 28.0 & 37.8 & 81.3 & 5.2 & 9.8 \\
    \bottomrule
    \end{tabular}
\end{table}

For SCDEE and PTPCG models, the scope of candidate arguments is narrowed by constructing communities or cliques in advance. Therefore, the training effect on small samples is reduced less than that of other baseline models, demonstrating the effectiveness of constructing entity-event corresponding relation. Eliminating entities unrelated to the current event is practicable because it relieves the model learning ability decline caused by the lack of training data. 

In summary, the model in this paper can handle the situations of single-event and multi-event and has excellent robustness on small samples. 
Furthermore, the features proposed herein are valuable in argument identification and play diverse roles.

\section{Related Work}\label{sec:related work}
Related studies on event extraction can be summarized into the following two aspects: sentence-level and document-level event extraction.

\textbf{Sentence-level Event Extraction.} Previous studies majorly focused on sentence-level event extraction \citep{lai2021graph,sheng2022cored,li2022dual,wan2022construction,lou2022translation,wan2023cfere}. \citet{Chen2015} designed a neural pipeline model to identify triggers first and then extract arguments. \citet{Nguyen2016} developed a joint model to extract event triggers and arguments based on recurrent neural networks. In addition, some work used graph neural networks to encode the syntactic dependency information and structures to extract events \citep{Nguyen2018,Sha2018,Liu2018,Yan2019,lai2020event,ahmad2021gate,Wan2021,wan2022multi}. 
In recent years, many other external characteristics and strategies have been adopted. For characteristics, the exploitation of open domain triggers, statistical information (e.g., token-event coreference frequency), saliency attribution, and pre-trained language models can be found in \citep{li2021treasures,liu2022saliency,tong2020improving,veyseh2021unleash}. Given strategies, \citet{Huang2020Biomedical} proposed a hierarchical policy network, and \citet{liao2021learning} argued a contrastive learning and mixspan strategy. \citet{Liu2020} and \citet{Du2020} tried to extract events in a machine reading comprehension and question-answer way. 
However, it is difficult for sentence-level models to extract multiple qualified events spanning across sentences \citep{xu2021document}.  

\textbf{Document-level Event Extraction.} 
With the corpus release for document-level event extraction, such as ChFinAnn \citep{Zheng2019} and RAMS \citep{ebner2020multi}, this task has attracted more and more attention recently \citep{lin2022cup,ma2022prompt}. \citet{ebner2020multi} designed a span-based argument linking model. A two-step method was proposed for argument linking by detecting cross-sentence arguments \citep{zhang2020two}. 
\citet{du2020document} tried to encode sentence information in a multi-granularity way, and \citet{li2021document} developed a neural event network model by conditional generation. \citet{ma2022prompt} and \citet{lin2022cup} exploited prompts and language models for document-level event argument extraction. Nevertheless, these studies only considered the sub-task of document-level event extraction (i.e., role filler extraction or argument extraction) and ignored the challenge of multi-events \citep{yang2021document}. 

Therefore, some other studies focused on the multi-event corpus (ChFinAnn). \citet{Yang2018} extracted events from a key-event sentence and found other arguments from neighboring sentences. \citet{Zheng2019} implemented event extraction following a pre-defined order of argument roles with an entity-based path expansion. Subsequently, \citet{xu2021document} built a heterogeneous interaction graph network to capture global interactions among different sentences and entity mentions. Their execution frameworks are based on \citet{Zheng2019}. \citet{yang2021document} extracted events in a parallel mode, overcoming the dependence on argument role order. 
\citet{huang2021exploring} and \citet{zhu2022efficient} took a different strategy. \citet{huang2021exploring} exploited sentence community to determine the corresponding relation of entity-event, and this was done with a maximal clique composed of pseudo-triggers in \citet{zhu2022efficient}. 

However, these methods are based on multiple independent sub-tasks and suffer from large computing complexity due to a large parameter size.
Therefore, this paper aims to develop a single-task extraction model for document-level multi-event and argument cross-sentence.

\section{Conclusion}\label{sec:conclusion}
This paper addresses the challenges of document-level event extraction by proposing a novel approach (TER-MCEE). We introduce the \textit{token-event-role} data structure, which captures the matching relations among tokens, events, and argument roles. This data structure allows us to unify the sub-tasks of entity extraction and multi-event extraction under recognized event types into a multi-classification problem for predicting argument role types in token-event pairs. By leveraging this structure, we can effectively capture the correlation semantics of tokens acting as arguments in different events.
To achieve comprehensive event extraction, we adopt a multi-channel mechanism that treats each event type as a separate channel. This approach eliminates the need for an event type judgment sub-task by clarifying the argument roles for a specific event type. Furthermore, we explore and incorporate several crucial features to enhance the extraction performance.
Extensive experiments are conducted on the ChFinAnn corpus to evaluate the effectiveness and robustness of our proposed scheme. The results demonstrate the superior performance of our approach. We provide in-depth analyses to explore the essential factors, including the handling of single and multiple events, ablation studies, and the performance on small samples, which significantly influence the event extraction performance.

\bibliographystyle{ACM-Reference-Format}
\bibliography{sample-base}


\end{document}